\documentclass[]{ceurart}

\usepackage{multirow}

\begin{document}

\copyrightyear{2023}
\copyrightclause{Copyright for this paper by its authors.
  Use permitted under Creative Commons License Attribution 4.0 International (CC BY 4.0).}

\conference{LM-KBC'23: Knowledge Base Construction from Pre-trained Language Models,
  Challenge at ISWC 2023}

\title{Expanding the Vocabulary of BERT for Knowledge Base Construction}

\author[1]{Dong Yang}[%
email=dong.yang@maastrichtuniversity.nl,orcid=0000-0003-4507-8162]

\author[1]{Xu Wang}[%
email=xu.wang@maastrichtuniversity.nl]

\author[1]{Remzi Celebi}[%
email=remzi.celebi@maastrichtuniversity.nl]

\address[1]{Institute of Data Science, Department of Advanced Computing Sciences, Maastricht University, The Netherlands}

\begin{abstract}
Knowledge base construction entails acquiring structured information to create a knowledge base of factual and relational data, facilitating question answering, information retrieval, and semantic understanding. The challenge called "Knowledge Base Construction from Pretrained Language Models" at International Semantic Web Conference 2023 defines tasks focused on constructing knowledge base using language model. Our focus was on Track 1 of the challenge, where the parameters are constrained to a maximum of 1 billion, and the inclusion of entity descriptions within the prompt is prohibited.

Although the masked language model offers sufficient flexibility to extend its vocabulary, it is not inherently designed for multi-token prediction. To address this, we present Vocabulary Expandable BERT for knowledge base construction, which expand the language model's vocabulary while preserving semantic embeddings for newly added words. We adopt task-specific re-pre-training on masked language model to further enhance the language model.

Through experimentation, the results show the effectiveness of our approaches. Our framework achieves F1 score of 0.323 on the hidden test set and 0.362 on the validation set, both data set is provided by the challenge. Notably, our framework adopts a lightweight language model (BERT-base, 0.13 billion parameters) and surpasses the model using prompts directly on large language model (Chatgpt-3, 175 billion parameters). Besides, Token-Recode achieves comparable performances as Re-pretrain. This research advances language understanding models by enabling the direct embedding of multi-token entities, signifying a substantial step forward in link prediction task in knowledge graph and metadata completion in data management.
  \footnote{Accepted for the workshop LM-KBC @ ISWC 2023. Our code and data are available at \url{https://github.com/MaastrichtU-IDS/LMKBC-2023}}
\end{abstract}

\maketitle

\section{Introduction}

Knowledge bases have a profound impact across diverse domains, offering transformative benefits. They enhance information retrieval systems, leading to increased efficiency and accuracy, thereby enabling users to swiftly locate relevant data \cite{nguyen_design_2022}. In the context of natural language processing, knowledge bases play a crucial role in elevating semantic comprehension and facilitating a range of language-related tasks \cite{zhang_dkplm_2022}. Moreover, these knowledge bases actively promote data integration and interoperability, making substantial contributions to the advancement of initiatives such as the Semantic Web and Linked Data \cite{bouaicha_semantic_2023}.
\begin{figure}
    \centering
    \includegraphics[width=1\linewidth]{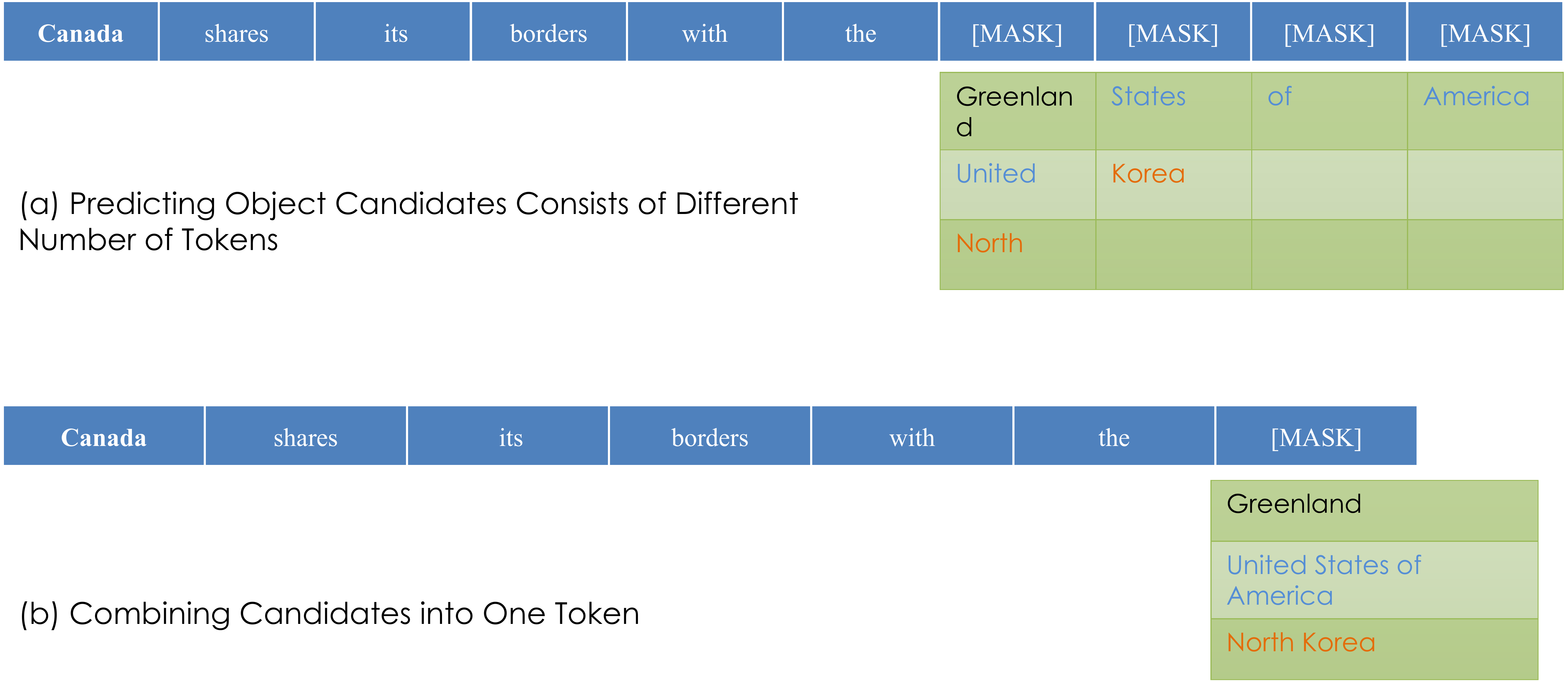}
    \caption{Predicting Objects Comprised of Multiple Lexical Tokens}
    \label{fig:multiple-token}
\end{figure}
In this work, we present our approach for the LM-KBC challenge \cite{singhania_lm-kbc_2023} at ISWC 2023, which focuses on knowledge base construction for 21 relations. The task of the challenge involves predicting objects based on given subject-relation pairs. For example, given the subject-relation pair  \textit{<Canada, CountryBordersCountry>}, the goal is to predict appropriate objects such as \textit{United States of America, Greenland}. In this challenge, each participant receives a set of subject-relation pairs and is tasked with identifying the appropriate objects for these pairs. Each subject-relation pair can be associated with zero, one, or multiple true objects, reflecting the complex nature of real-world scenarios.

We participate in \textbf{track 1} of the challenge \cite{singhania_lm-kbc_2023}, where the parameters of language model is limited up to 1 billion.  We selected the BERT \cite{devlin_bert_2019} model as the encoder and performed the Filled-Mask task to retrieve object candidates \cite{li_task-specific_2022}. For each [mask] token, the language model independently assigns a confidence score to all tokens within its vocabulary. However, the original filled-mask task was designed to select the best single candidate, rather than multiple top candidates. Furthermore, the target object entity may consist of multiple tokens, and for a given subject-relationship pair, the number of potential objects can vary. The original model for filled-mask tasks is not inherently formulated to predict multiple objects comprised of numerous tokens. For example (Figure \ref{fig:multiple-token} (a)),  the correct answers of given subject-relation pair \textit{<Canada, CountryBordersCountry>} are combinations of tokens, such as \textit{('Greenland'), ('United', 'States', 'of', 'America')}. However, extracting entities from the filled-mask task is not a straightforward process, as it presents an permutation problem due to the language model's independent prediction of each token.

To be able to predict multiple candidates with multiple tokens, we modify the token embedding layer and the output embedding layer of BERT. We expand the vocabulary of language model (e.g. BERT), and the object composed of multiple tokens (e.g. "United States of America") is treated as a distinct token. As shown in Figure \ref{fig:multiple-token} (b), our approach entails grouping token combinations into single tokens respectively. However, a drawback of this method is that the newly created tokens cannot leverage the information provided by the language model. 


To address this challenge, we propose Vocabulary Expandable BERT (VE-BERT), which aims to provide an initial semantic vector for newly added entities. We build a vocabulary by querying entities on WikiData using the predicates defined by the challenge. We conduct experiments to evaluate the effectiveness of TR, and the results demonstrate an obvious improvement in the F1 score. This indicates that leveraging the Token Re-code task enhances the model's ability to align newly defined entities with their constituent tokens, thereby providing a more accurate semantic representation for newly added entities than randomly initialized embeddings. And compare TR with re-pretrain the model with additional raw text, TR do not need extra training and achieves same improvement.

We collect sentences from wikipedia, filter sentences according to the frequency of the entities in our vocabulary.  Experiments shows the task-specific pre--train is effective.  We also categorised the entities and find the best threshold of each predicates on valid set.

The main contributions of this paper is:
\begin{itemize}  
\item Proposing a method to expand the vocabulary of a language model (E.g. BERT)  while preserving the semantic meaning of the newly added entities. 
\item Conducting experiments to verify the effectiveness of re-pretrain on raw text for knowledge base construction.
 \end{itemize}

We achieved 0.362 at validation set and 0.323 on hidden test set with bert-base-cased model, which is a relative light language model,  with only 130 million parameters.

\section{Related Work}

\subsection{Masked Language Model}
The Bidirectional Encoder Representations from Transformers (BERT) \cite{devlin_bert_2019} model has transformed the field of natural language processing (NLP) since its introduction. BERT's primary contribution lies in its ability to generate contextualized word embeddings, capturing bidirectional context information, and producing rich semantic representations. By pre-training on large-scale unlabeled text using a masked language modeling objective, BERT learns a deep representation of language structures, enabling it to capture complex linguistic patterns and relationships.  The BERT (Bidirectional Encoder Representations from Transformers) \cite{devlin_bert_2019} model uses two fundamental types of embeddings: input embeddings and output embeddings, each of which serves a distinct yet interrelated role in the model's functioning.  Given an input sequence of tokens $X = (x_1, x_2, ..., x_n)$, where each $x_i$ represents a token (word or sub-word) in the sequence. The input tokens are first transformed into embeddings $E = (e_1, e_2, ..., e_n)$, where each $e_i$ is the embedding representation of the token $x_i$. These embeddings are then processed through multiple layers of Transformer architecture.
\begin{equation}
    P(x_i | e_1, e_2, ..., e_n) = T(E) \times O
\end{equation}

Where $P(x_i | e_1, e_2, ..., e_n)$ is the predicted probability distribution over the vocabulary for the i-th position, conditioned on the embedding of all tokens in the sequence. The $E$  refers to the token embeddings of the input tokens $X$. The $T$ refers to stacked transformer layers. $O \in R^{l \times v}$, where l is the length of the width of last hidden layer of the stacked transformer layers $T$, v is the number of the vocabulary.

\textbf{The input embeddings}  capture the inherent semantic and contextual information of the input tokens. Specifically, BERT breaks down words into sub-word units (sub-tokens). These sub-word embeddings are then combined with positional embeddings to encode both the content and the position of the tokens within the input sequence. The input embeddings go through a series of transformations as they pass through BERT's layers. Initially, these embeddings are fed into the model's self-attention mechanism, which enables the model to capture contextual relationships between tokens in both directions (left-to-right and right-to-left) in the input sequence. This bidirectional context is a significant departure from previous models that relied solely on left-to-right or right-to-left information flow.  \textbf{The output embeddings} refer to the representations of the tokens that are obtained after the input embeddings have been processed through BERT's layers. These output embeddings encapsulate the model's learned understanding of the input text's semantics and context. Output embeddings can be utilized for various downstream tasks, such as text classification, named entity recognition, question answering, etc. 

The XLNet \cite{yang_xlnet_2019} and GPT (Generative Pre-trained Transformer) \cite{brown_language_2020} models are two other prominent advancements in the field of natural language processing (NLP). Both models have made use of transformers and self-supervised learning techniques, leading to significant contributions to language understanding and generation tasks. Gururangan \cite{gururangan_dont_2020} built separate pretrained models for specific domains with a universal language model ROBERTA on four domains (biomedical and computer science publications, news and reviews) and eight classification tasks (two in each domain). Their experiments showed that continued pre-training with additional corpus on the domain consistently improves performance on tasks from the target domain, in both high- and low-resource settings.

\subsection{Knowledge Base Construction}
Kalo   \cite{pan_knowlybert_2020} introduced a composite query-answering architecture that integrates knowledge graphs with the BERT masked language model to augment the precision of query outcomes. This approach fuses the structural and semantic attributes of knowledge graphs with the textual knowledge from language models. The model uses multiple MASK tokens to predict tokens. And then find the most likely entities from the combinations of the individual predictions.  Li \cite{li_task-specific_2022} is the winner of the LM-KBC 2023. They proposed a model based on BERT-large-cased, to improve performance in the following three aspects: (1) LM representation of masked object tokens;(2) entity generator; (3) candidate object selection.  The author used additional triples to train the language model and therefore, leading significant improvement. The skills related to prompting can be categorized into four types: (1) incorporating type information to entities; (2) simplifying and condensing prompts; (3) generating prompts by extracting relevant sentences from Wikipedia; (4) selecting different prompts for the same relation based on the type of entity. Besides, they remove pronouns and determiners from the candidates and find the optimal threshold of each object-relation pair and use the original score of the predictions rather than softmax.

\section{Model}
\begin{figure}
    \centering
    \includegraphics[width=1\linewidth]{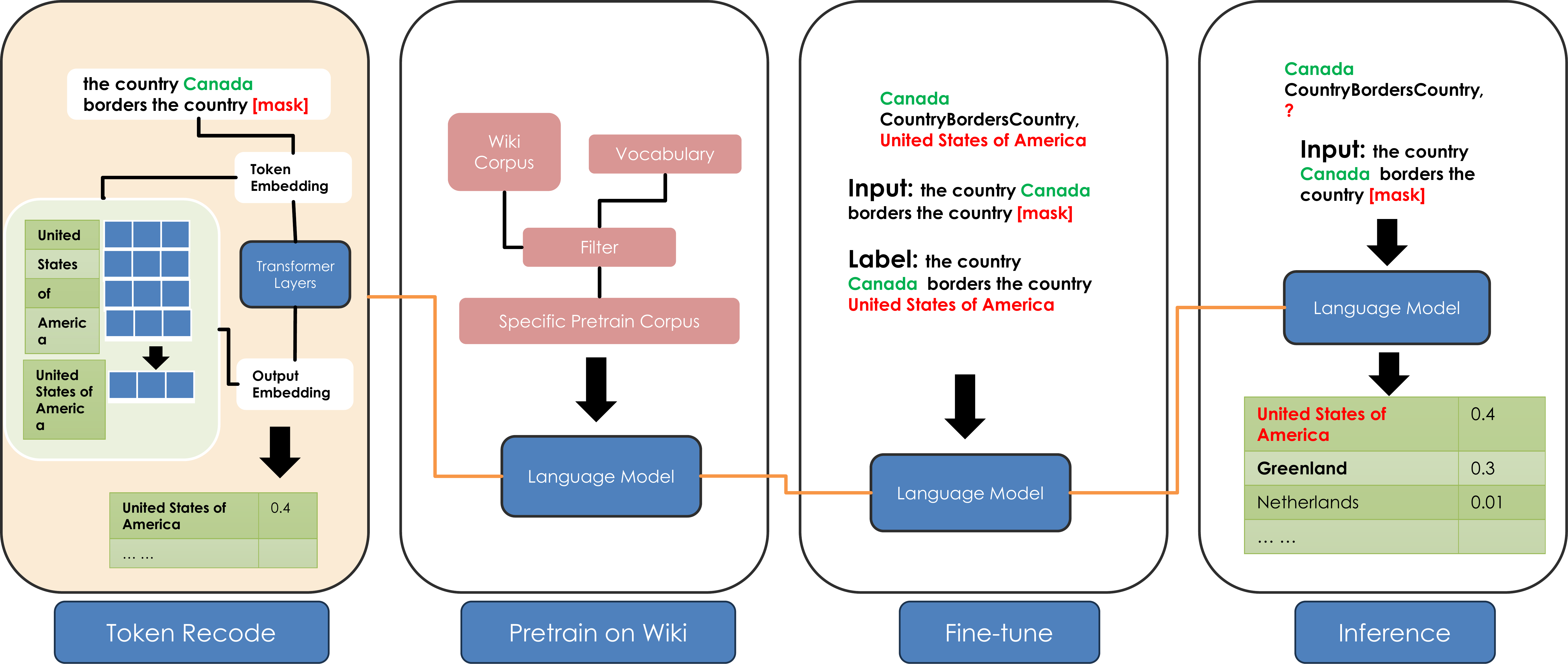}
    \caption{The Structure of Our Framework}
    \label{fig:overall}
\end{figure}

In Figure \ref{fig:overall}, we provide an overview of our framework. 
We introduce the Token Recode method, which modify the token embedding layer and output embedding layer, to be able to encode and predict the object entities with multiple tokens. The modified BERT is named the Vocabulary Expandable BERT (VE-BERT). In Token Recode layer, we first initialize the token embeddings and output embeddings for multi-token entities in a well-known Language Model, specifically "bert-base-cased" as referenced within this work.  We re-pretrain the VE-BERT based filled-masked model, using a corpus sourced from the Wikipedia pages, called WikiCorpus. Sentences are chosen for pretraining based on how frequently the entities from our WikiData-derived vocabulary appear together in the sentence. We describe the details about the vocabulary generation in the following subsection. 

We then fine-tune this filled-masked model using the training set provided within the challenge and use the model to predict the object candidates for the validation set and test set. In Inference step, we select the best thresholds for relation types based on the performance on the validation set, and use the selected thresholds to predict the object candidates for the test set.  In addition, for the relations "PersonHasNumberOfChildren" and "SeriesHasNumberOfEpisodes", which have numbers in their range, we check whether the resulting candidates for an object is a number. Finally, we disambiguate the predicted entities by WikiData-API.

\subsection{Vocabulary}
We categorized the entity by their roles in a triple (subject or object) as shown in Table \ref{tab-schema}. 
\begin{table}
\caption{The entity type of relation}\label{tab-schema}
\begin{tabular}{|c|c|c|}
\hline
Relation & Subject Type & Object Type \\
\hline
BandHasMember & Band & Person \\
CityLocatedAtRiver & City & River  \\
CompanyHasParentOrganisation & Company & Company \\
CompoundHasParts & Compound & Part\\
CountryBordersCountry & Country & Country\\
CountryHasOfficialLanguage & Country & Language\\
CountryHasStates & Country & State\\
FootballerPlaysPosition & Person & Position\\
PersonCauseOfDeath & Person & Cause\\
PersonHasAutobiography & Person & Autobiography\\
PersonHasEmployer & Person & Company\\
PersonHasNoblePrize & Person & Prize\\
PersonHasNumberOfChildren & Person & Number\\
PersonHasPlaceOfDeath & Person & City\\
PersonHasProfession & Person & Profession\\
PersonHasSpouse & Person & Person\\
PersonPlaysInstrument & Person & Instrument\\
PersonSpeaksLanguage & Person & Language\\
RiverBasinsCountry & River & Country\\
SeriesHasNumberOfEpisodes & Series & Number\\
StateBordersState & State & State\\
\hline
\end{tabular}
\end{table}

We created a task-specific vocabulary using entities from the challenge dataset (i.e., training, validation and test set ). Additionally, we extended the entity set from WikiData Knowledge Graph.  We queried  all the entities that has at least one relation listed in the challenge dataset. Table \ref{tab-vocabulary} gives the  number of collected entities for each type in the vocabulary. 
\begin{table}
\caption{The count of entities for each type in the vocabulary}\label{tab-vocabulary}

\begin{tabular}{|c|c|}
\hline
Entity Type & Number \\
\hline
Company & 1654 \\
Country & 248 \\
Language & 252 \\
Number & 1063 \\
City & 111 \\
Profession & 233 \\
Instrument & 58 \\
State & 5470 \\
Person & 2287 \\
River & 1999 \\
Position & 78 \\
Cause & 78 \\
Autobiography & 375 \\
Part & 407 \\
Prize & 7 \\
\hline
Total & 14320\\
\hline
\end{tabular}
\end{table}

 The BERT model is primarily designed for a "filled-mask" or "masked language modeling" task, where certain tokens in a sentence are masked, and the model is trained to predict the original tokens. The objective is to learn contextualized representations of words that take into account their surrounding context.

\subsection{Token Recode }

We introduce modifications to the token embedding and output embedding components of the BERT model, enabling the generation of embeddings for newly introduced phrases based on their constituent tokens. For instance, consider the entity 'United States of America', which is partitioned into individual tokens as ['United', 'States', 'of', 'America']. The token and output embeddings for the phrase 'United States of America' are computed as the average of the respective embeddings for its tokens ['United', 'States', 'of', 'America'].


In a more general context, we denote the newly introduced phrases as $W$, with its associated tokens represented as $S=(s_1, s_2, s_3, \ldots, s_n)$. The initial token embeddings for these tokens are denoted as $t=(t_1, t_2, t_3, \ldots, t_i, \ldots, t_n)$, while the original output embeddings are denoted as $o=(o_1, o_2, o_3, \ldots, o_i, \ldots, o_n)$. Subsequently, the new token embedding for word $W$ is denoted as $T$, and the new output embedding is denoted as $O$.
Then we obtain the token embedding and output embedding of a word $W$ by averaging the original token embedding $t$ and output embedding $o$ of its corresponding tokens $t$:

\begin{equation}
T =  \frac{1}{n}\sum_{i=1}^n t_i 
\end{equation}

\begin{equation}
O =  \frac{1}{n}\sum_{i=1}^n o_i 
\end{equation}

Then we normalize the new token embedding $T$ and new output embedding $O$ of word $W$. Where $T \in R^{n}, O \in R^{n} $, The subscript $j$ serves as the index corresponding to a particular embedding vector. 
\begin{equation}
T =  \frac{T_j}{\sqrt{\sum_{j=1}^n T_j ^ 2}}
\end{equation}

\begin{equation}
O =  \frac{O_j}{\sqrt{\sum_{j=1}^n O_j ^ 2}}
\end{equation}

\subsection{Pre-training on Wikipedia }
We generate the embedding of a word of BERT model by deriving the embedding of its constituent tokens. Furthermore, we pre-train the model by conducting filled-mask task on our collected Wikipedia corpus. The sentence in our corpus is selected based on the criterion that the sentences include the entities listed in our vocabulary. Table \ref{tab-corpus} indicates the number of sentence that include specific entity type. Please note that a sentence can contain multiple entities, resulting in the cumulative count of sentences for each entity type being greater than the actual overall sentence count.

\begin{table}
\caption{The quantity of sentences for each entity type}\label{tab-corpus}
\begin{tabular}{|c|c|}
\hline
Entity Type & Quantity of Sentences \\
\hline
Person & 22114 \\
Country & 10845 \\
Series & 3665 \\
State & 36076 \\
Company & 12981 \\
Band & 5246 \\
River & 9770 \\
Autobiography & 879 \\
City & 10015 \\
Compound & 408 \\
Profession & 5062 \\
Part & 649 \\
Position & 444 \\
Cause & 1041 \\
Language & 3732 \\
Instrument & 932 \\
Prize & 576 \\
\hline
Total & 51496\\
\hline
\end{tabular}
\end{table}

\subsection{Fine-tune on knowledge base construction}
\label{sec-fine-tune}
Contemporary language models, exemplified by BERT \cite{devlin_bert_2019}, have undergone extensive training on a corpus of diverse textual data at a significant scale. This inherent capacity for comprehensiveness suggests that a process of "rekindling" the models' awareness of the specific categories of information they are expected to recall during fine-tuning could potentially yield performance enhancements. In congruence with this perspective,  \cite{li_task-specific_2022} have empirically demonstrated the utility of fine-tuning in augmenting the performance of language models in knowledge base construction.

The process of fine-tuning for the knowledge bases construction can be summarized as follows: given a subject-relation-object triple, this triple is transformed into a coherent sentence using a corresponding prompt template.  Relevant tokens related to the object entity are hidden within the sentence. The subsequent task involves training BERT models using the masked sentence as input, aiming to effectively uncover the hidden tokens.

\section{Experiments}
We use transformers on PyTorch to build our system, conducting experiments on V100 32G. For pre-training on wiki sentence task, the learning-rate is set to  $2e^{-5}$ and epoch numbers is 20 . For fine-tune task, the learning-rate is set to $2e^{-5}$ and the number of epoch is 5. We adopt the disambiguation function released by the challenge. But for the predicates of which the type of object entities is "Number", we used the predicted numbers directly.  The code is available at \href{https://github.com/MaastrichtU-IDS/LMKBC-2023}{https://github.com/MaastrichtU-IDS/LMKBC-2023}  

\section{Results}
\begin{table}

\centering

\caption{Results on validation set }
\label{tab-method}
\begin{tabular}{|c|c|c|c|c|}
\hline
Method  &   Parameter quantity (billion)&Precision   &   Recall  & F1 \\
\hline
$prompt-directly_{bert-base-cased}$&  0.11&0.131& 0.474 & 0.112 \\
 $prompt-directly_{bert-large-cased}$& 0.345& 0.368& 0.161&0.142\\
$prompt-directly_{facebook/opt-1.3b}$&  1.3&0.073 & 0.101 &  0.039  \\
 $prompt-directly_{GPT3-ner}$&  175&0.308& 0.210&0.218\\
\hline
$baseline_{bert-base-cased}$&   \multirow{4}{*}{ 
 0.13 }  &0.442 & 0.382 &  0.311 \\
$token-recode$&  & 0.426 &  0.400 & 0.334 \\
 $re-pretrain$&  &0.440& 0.405&0.331\\
$VE-BERT_{token-recode+re-pretrain}$&     &0.493     &0.443   &  0.362  \\
\hline
\end{tabular}
\end{table}

We conducted experiments to evaluate the performance of different methods for knowledge base construction. Table~\ref{tab-method} summarizes the results of our experiments. $prompt-directly$  models initialize prompt templates using subject-predicate pairs and subsequently predict missing object entities. These models and hyperparameters is provided by the challenge LM-KBC 2023 \cite{singhania_lm-kbc_2023}.


 The $token-recode$ method successfully improves extracting correct object entities, achieves 2 percentages improvement.   Re-pretraining on Wikipedia sentences improves constructing knowledge bases by almost 2 percentages. The performance of the method $token-recode$ is comparable to the method $re-pretrain$, without additional computation. 

Our final model, Vocabulary Expandable BERT (VE-BERT) achieves nearly 5 percentages improvement, due to the combination of the methods $token-recode$ and $re-pretrain$. This shows that $token-recode$ method can provide a high quality initial embeddings for newly added entities, and thus help the process of re-pretrain.

In our final experiment, the Vocabulary Expandable BERT (VE-BERT) model shows a nearly 5\% improvement. This increase is due to the combined use of two methods:  token-recode and re-pretrain. Interestingly, this combined improvement is greater than the sum of improvements from using each method separately. This suggests that the token-recode  method offers high-quality initial embeddings for new entities, making the re-pretrain process more effective.

\begin{table}[h!]
    \centering
    \caption{The results of baseline and ours system on validation set }  \label{tab-detail}
    \begin{tabular}{|c|c|c|c|c|c|c|}
        \hline
        \multirow{2}{*}{Relation} & \multicolumn{3}{|c|}{$prompt-directly_{bert-base-cased}$} & \multicolumn{3}{|c|}{$VE-BERT_{bert-base-cased}$} \\\cline{2-7}
         & Precision & Recall & F1 & Precision & Recall & F1 \\\hline
BandHasMember                &  0.000& 0.960 & 0.000  &      0.559  &  0.055  &  0.057    \\  
CityLocatedAtRiver           &  0.018& 0.145 & 0.020  &      0.558  &  0.200  &  0.182    \\  
CompanyHasParentOrganisation &  0.060& 0.640 & 0.053  &      0.675  &  0.610  &  0.577    \\  
CompoundHasParts             &  0.164& 0.475 & 0.179  &      0.714  &  0.863  &  0.760    \\  
CountryBordersCountry        &  0.372& 0.709 & 0.443  &      0.495  &  0.485  &  0.452    \\  
CountryHasOfficialLanguage   &  0.760& 0.685 & 0.668  &      0.838  &  0.691  &  0.708    \\  
CountryHasStates             &  0.002& 0.076 & 0.005  &      0.221  &  0.198  &  0.195    \\  
FootballerPlaysPosition      &  0.413& 0.143 & 0.206  &      0.178  &  0.705  &  0.278    \\  
PersonCauseOfDeath           &  0.040& 0.012 & 0.019  &      0.680  &  0.680  &  0.680    \\  
PersonHasAutobiography       &  0.000& 0.950 & 0.000  &      0.890  &  0.010  &  0.010    \\  
PersonHasEmployer            &  0.000& 0.970 & 0.000  &      0.345  &  0.028  &  0.028    \\ 
PersonHasNoblePrize          &  0.000& 0.000 & 0.000  &      0.532  &  0.755  &  0.593    \\  
PersonHasNumberOfChildren    &  0.000& 0.000 & 0.000  &      0.282  &  0.645  &  0.388    \\  
PersonHasPlaceOfDeath        &  0.091& 0.641 & 0.088  &      0.561  &  0.556  &  0.512    \\ 
PersonHasProfession          &  0.000& 0.700 & 0.000  &      0.135  &  0.197  &  0.145    \\  
PersonHasSpouse              &  0.000& 0.890 & 0.000  &      0.720  &  0.020  &  0.020    \\  
PersonPlaysInstrument        &  0.000& 0.110 & 0.000  &      0.493  &  0.795  &  0.575    \\ 
PersonSpeaksLanguage         &  0.383& 0.310 & 0.316  &      0.731  &  0.662  &  0.654    \\  
RiverBasinsCountry           &  0.423& 0.402 & 0.345  &      0.516  &  0.512  &  0.469    \\  
SeriesHasNumberOfEpisodes    &  0.000& 1.000 & 0.000  &      0.127  &  0.380  &  0.189    \\  
StateBordersState            &  0.016& 0.132 & 0.016  &      0.097  &  0.250  &  0.125    \\  
\hline
Average                      &  0.131& 0.474 & 0.112  &      0.493  &  0.443  &  0.362    \\

\hline
    \end{tabular}
    \label{tab:final_experiment}
\end{table}

Table~\ref{tab-detail} shows the final results for our model VE-BERT on the  valid set of the challenge LM-KBC 2023 \cite{singhania_lm-kbc_2023}.
The result of VE-BERT demonstrates proficiency in predicting general knowledge predicates, such as \textit{CompoundHasParts} and \textit{CountryHasOfficialLanguage}. However, it exhibits limitations in predicting privacy information, exemplified by predicates like \textit{PersonHasAutobiography} and \textit{PersonHasSpouse}. Additionally, VE-BERT is constrained to predicting multi-token entities present within its vocabulary. Owing to a relatively low occurrence of \textit{Name} entities within the vocabulary, the framework underperforms in predicting predicates where the object type is \textit{Name}, such as in \textit{PersonHasSpouse} and \textit{BandHasMember}.

\section{Conclusions}

Our research aims to enhance the construction of knowledge bases through the employment of lightweight Language Models, adhering to the constraints delineated in Track 1, which restricts the model parameters to one billion. We introduce the Vocabulary Expandable BERT (VE-BERT), a modification of the standard BERT architecture that involves alterations to both the input and output embedding layers. The model is further enriched by pre-training on a task-specific corpus and subsequent fine-tuning on a training set. Experimental results validate the efficacy of our proposed token-recode method, which exhibits augmented performance when coupled with a re-pretraining task.

As we navigate the complexities of factual statement extraction from language models, we identify two pivotal areas warranting future investigation: the application of VE-BERT in link-prediction tasks within knowledge graphs and in missing value prediction within data management.

\begin{acknowledgments}
  The authors thank the challenge organizers for their timely and helpful response to inquiries, and the reviewers for their valuable comments. This work is supported by China Scholarship Council (202207010004). 
\end{acknowledgments}

\bibliography{refs}

\appendix

\end{document}